# Spatio-Temporal Deep Learning-Assisted Reduced Security-Constrained Unit Commitment

Arun Venkatesh Ramesh, *Member, IEEE* and Xingpeng Li, *Senior Member, IEEE*

*Abstract*— Security-constrained unit commitment (SCUC) is a computationally complex process utilized in power system day-ahead scheduling and market clearing. SCUC is run daily and requires state-of-the-art algorithms to speed up the process. The constraints and data associated with SCUC are both geographically and temporally correlated to ensure reliability of the solution, which further increases the complexity. In this paper, an advanced machine learning (ML) model is used to study the patterns in power system historical data, which inherently considers both spatial and temporal (ST) correlations in constraints. The ST-correlated ML model is trained to understand spatial correlation by considering graph neural networks (GNN) whereas temporal sequences are studied using long short-term memory (LSTM) networks. The proposed approach is validated on several test systems namely, IEEE 24-Bus system, IEEE-73 Bus system, IEEE 118-Bus system, and synthetic South-Carolina (SC) 500-Bus system. Moreover, B-θ and power transfer distribution factor (PTDF) based SCUC formulations were considered in this research. Simulation results demonstrate that the ST approach can effectively predict generator commitment schedule and classify critical and non-critical lines in the system which are utilized for model reduction of SCUC to obtain computational enhancement without loss in solution quality.

## Nomenclature

Indices:

| | |
|---|---|
| $g$ | Generator index. |
| $k$ | Line index. |
| $t$ | Time period index. |
| $n$ | Bus index. |
| $m$ | Data sample index. |

Sets:

| | |
|---|---|
| $E$ | Set of edges in a graph. |
| $G$ | Set of generators. |
| $g(n)$ | Set of generators connecting bus $n$. |
| $K$ | Set of all transmission elements. |
| $N$ | Set of all buses. |
| $T$ | Set of Time intervals. |
| $\delta^+(n)$ | Set of lines with bus $n$ as receiving bus. |
| $\delta^-(n)$ | Set of lines with bus $n$ as sending bus. |
| $M$ | Total samples. |
| $M^{test}$ | Test Samples. |
| $M^{train}$ | Training Samples. |
| $M^{val}$ | Validation Samples. |
| $V$ | Set of vertices or nodes in a graph. |

Parameters:

| | |
|---|---|
| $A$ | Adjacency matrix describing the graph connection. |
| $b_k$ | Susceptance of line $k$. |
| $c_g$ | Linear cost for generator $g$. |
| $c_g^{NL}$ | No-load cost for generator $g$. |
| $c_g^{SU}$ | Start-up cost for generator $g$. |
| $d_{n,t}^m$ | Demand of bus $n$ in time period $t$ for sample $m$. |
| $d_{n,t}^{ini}$ | Initial nodal demand of bus $n$ in time period $t$. |
| $EF$ | Edge features matrix. |
| $NF$ | Node feature matrix. |
| $P_g^{max}$ | Maximum capacity limit of generator $g$. |
| $P_g^{min}$ | Minimum power limit of generator $g$. |
| $P_k^{max}$ | Long-term thermal limit for line $k$. |
| $R_g^{10}$ | 10-minute outage ramping limit of generator $g$. |
| $R_g^{hr}$ | Regular hourly ramping limit of generator $g$. |
| $R_g^{SD}$ | Shut-down ramping limit of generator $g$. |
| $R_g^{SU}$ | Start-up ramping limit of generator $g$. |
| $Y_N$ | True labels for generator commitment schedule. |
| $Y_E$ | True labels for critical and non-critical lines. |

Variables:

| | |
|---|---|
| $P_{g,t}$ | Output of generator $g$ in time period $t$. |
| $P_{k,t}$ | Lineflow of line $k$ in time period $t$. |
| $r_{g,t}$ | Reserve of generator $g$ in time period $t$. |
| $u_{g,t}^m$ | Commitment status of generator $g$ in time period $t$ for generated sample $m$. |
| $v_{g,t}$ | Start-up variable of generator $g$ in time period $t$. |
| $\theta_{i,t}$ | Phase angle of bus $i$ in time period $t$. |
| $\theta_{o,t}$ | Phase angle of bus $o$ in time period $t$. |
| $PTDF_{k,n}^{ref}$ | Sensitivity factor for power flow in line $k$ due to injection in bus $n$ withdrawn from the slack bus *ref*. |

## I. Introduction

The operations of power grid is a complex process in any vertically integrated and competitive markets. Typically, this involves a computationally rigorous process of solving a security-constrained unit-commitment (SCUC) problem which is a mixed-integer linear programming (MILP) problem. The SCUC problem is already known to be a hard problem to solve given the characteristics of identifying the commitment schedule and dispatch of generators to meet a forecasted load. Typically, in USA, the independent system operators are responsible for collecting the generator and load bids and clearing the day-ahead markets which is financially binding with the grid operation of the next day [1].

Since reliability is a crucial component, the unit-commitment is modelled with security constraints which characterize the physical requirements of the system. These define requirements for network topology, thermal limits and

Arun Venkatesh Ramesh and Xingpeng Li are with the Department of Electrical and Computer Engineering, University of Houston, Houston, TX, 77204, USA. (e-mail: aramesh4@uh.edu; xingpeng.li@asu.edu).

restrictions of generator capacity limits, ramp limits and characteristics which implies that the resulting base solution is then security constrained. The goal of the SCUC is to provide the lowest cost of operations of resources to meet the demand, [2]-[6].

However, such restrictions, especially the ON/OFF scheduling of the generators, make the problem computationally intensive for large systems whereas the ISOs' have limited time to post an optimal and feasible solution for the day-ahead market which also maintains the reliability [7]. Therefore, advancements are required to provide significant time saving benefits without loss in solution quality effectively to solve the SCUC problem. Several research proposes the use of heuristics and/or decomposition algorithms for obtaining time-saving benefits [8]-[9]. Techniques involving machine learning (ML) have recently gained popularity in energy market operations and planning through promising results by offline trained models which can be utilized in an online setting [10]-[11].

ML as a tool offers several benefits if there is historical information to learn patterns that aid in decision-making support. Learning-based methods are predominantly utilized in SCUC to reduce complexity and increase computational ability. In particular, in SCUC, four main approaches are noticed for reducing problem complexity: (i) replacing the optimization with ML [12]-[13], (ii) reducing non-critical constraints [14]-[18], (iii) warm-start solutions [19], and (iv) reducing variables [20]-[22]. Replacing optimizations with ML predictions is not the right approach since this does not guarantee a feasible, optimal or a secured solution as noted in [19], [21]-[22]. However, to our understanding a combination of warm-start, constraint reduction and variable elimination have not been implemented concurrently.

Though ML is implemented in prior research, prior research only focusses on the conceptual execution with basic/vanilla ML architectures [12]-[22]. One way to ensure that information is well-realized is through deep learning with multiple layers in deep neural networks (DNN). But in reality, all relationships in the SCUC problem are not effectively realized with DNN and may require additional steps that are time-consuming to ensure feasibility of learning-based solution when used in an optimization process, [23]. To reduce such additional steps, advanced ML architectures that leverage the temporal correlation of the data and the spatial correlation in the networks can be used together to improve the ML predictions.

SCUC model is temporally correlated since the actions in earlier periods can be related to following periods which implies it is important to understand the sequence patterns in the data. For temporal correlation, recurrent neural networks (RNN) and long short-term memory networks (LSTM) are popularly utilized to study sequences in data [24]-[25]. LSTM has advantage over RNN to avoid vanishing gradients which avoids training errors [26]. LSTM performs well for prediction related tasks using time-series data [25]-[26].

Typically, most research works on flat/grid data as seen in [10]-[26] but power system network involves spatial/geographical correlation in addition to temporal correlation. This negates the topographic information in each snapshot of the data and relies on ML to learn such relationships which can be too optimistic. To overcome this issue, graph-based approaches such as graph neural networks (GNN) can be useful as seen in [27]-[28]. Topology aware ML models in power system have been studied in economic dispatch in [29] and [30]. In [29], a convolutional neural network (CNN) based model was utilized to learn hotspots using color-coded 2-D grid data, but this does not utilize the special/geographical features completely. The GNN model utilized in [30] predicts critical lines in the systems to reduce computations in economic dispatch whereas in [31], it was utilized to reduce computational complexity in reliability assessment or reliability unit commitment (RUC). However, GNN models have not been studied in SCUC problems yet.

Additionally, we noticed that the combination of both GNN and LSTM model to realize a spatio-temporal (ST) correlation has not been implemented in power systems particularly in solving SCUC to the best of our knowledge. An ST-based ML model can bring several benefits as seen in [32]-[33]. In [32], a CNN-LSTM ST model was utilized in weather data; and in [33] GNN ST model was utilized to study traffic data. But the traffic network does not have an Ohm's law congestion effect which makes the power system network unique and requires specific graph features to be modelled and trained. The combination of GNN and LSTM architectures would provide the ST awareness to learn relationship between inputs and outputs and benefit the power system in various applications.

Hence, in this paper an advanced supervised ML architecture is proposed which utilizes GNN and LSTM layers to consider a ST correlation of input and output data. Since the GNN layer is utilized, each data sample is created as a graph. The use of GNN layer also enables us to perform a node level and edge level classification. This implies that two ML models can be trained to study both generator $g$ commitment schedule and line $k$ criticality, respectively, in relation to the nodal demand profiles which can be later utilized to predict concurrently. Therefore, for both ML models, the graph inputs include node features that have a 24-hour nodal/bus demand for each period, $t$, and edge features that consider thermal limits and line properties for each line $k$. The first ML model performs a ST node classification (NC) where the graph outputs are each generator $g$ commitment schedule in each time period $t$. The second ML model is an edge classification (EC) model with graph outputs as line flags for each line $k$ in each time period $t$. The commitment status of one implies the generator is ON whereas zero represents the generator is OFF. Similarly, line congestion status of one implies the branch is critical and zero implies branch is non-critical. The predictions of ML models are post-processed using confidence-based (probability threshold) variable and constraint reduction in SCUC to obtain a reduced-SCUC (R-SCUC) process. The contributions of this paper are presented as follows:
- An offline advanced ST ML architecture using GNN and LSTM layers for NC is proposed to predict generation commitment schedule.
- An offline GNN-based model is developed to perform EC using nodal features as demand profiles and edge features as line properties to predict line criticality.
- A graph generation process is implemented to convert power system data to graph-based snapshots which can be utilized for GNN layers.

- The proposed ST-R-SCUC process which performs variable and constraint reduction can significantly improve the computational efficiency without loss in solution quality.

The rest of this paper is organized as follows. Section II presents the B-θ and power transfer distribution factor (PTDF) based formulations for SCUC model. Section III describes the graph generation procedure while Section IV details the offline step including ML architecture, training, confidence metrics, post-process procedure and the proposed methods while the results are discussed in Section V. Finally, Section VI concludes the paper.

## II. SCUC FORMULATION

In this paper, two formulations of SCUC are utilized, namely, B-θ formulation and PTDF formulation. The formulation only differs in line flow calculation whereas the other constraints are the same. It can be noted that both B-θ formulation and PTDF formulation will result in the same solution, but PTDF formulation inherently has reduced continuous variables and constraints compared to B-θ formulation.

The objective of SCUC is to minimize the operational cost of generators, which includes the production, start-up and no-load costs (1). This is optimized subject to power flow, generation and nodal balance requirements. The physical requirements of generating resources are modeled in (2)-(11). Generators are restricted to their minimum and maximum production limits enforced by (2) and (3) respectively. Reserves are ramp restricted and are allocated to handle the loss of the largest generation unit in the system in under 10-min by (4) and (5), respectively. The upward and downward ramping of generation units restricted for each period or hour in (6) and (7), respectively. Each generator has a different restriction for up-time and down-time and is modeled through (8)-(10). Finally, the generator commitment status and start-up variables are binary variables as shown in (11).

The base-case physical power flow constraint is represented through (12)-(14). Power flows can be calculated with either B-θ formulation (12), or PTDF formulation (13). The physical thermal limits of transmission elements are modelled in (14). Supply should always meet the demand which is modelled by the nodal power balance in (15). Slack bus is consistently modelled for both formulations. For PTDF formulation the sensitivity factors were calculated using slack bus, *ref*. In B-θ formulation, slack is modelled by setting phase angle of bus *ref* as zero in (16).

*Objective:*
$$\text{Min } \sum_g \sum_t (c_g^{NL} u_{g,t} + c_g^{SU} v_{g,t} + c_g P_{g,t}) \quad (1)$$
*s.t.:*
*Generation constraints:*
$$P_g^{min} u_{g,t}^m \leq P_{g,t}, \forall g \in G, t \in T \quad (2)$$
$$P_{g,t} + r_{g,t} \leq P_g^{max} u_{g,t}^m, \forall g \in G, t \in T \quad (3)$$
$$0 \leq r_{g,t} \leq R_g^{10} u_{g,t}^m, \forall g \in G, t \in T \quad (4)$$
$$\sum_{q \in G} r_{q,t} \geq P_{g,t} + r_{g,t}, \forall g \in G, t \in T \quad (5)$$
$$P_{g,t} - P_{g,t-1} \leq R_g^{hr} u_{g,t-1}^m + R_g^{SU} v_{g,t}, \forall g \in G, t \in T \quad (6)$$
$$P_{g,t-1} - P_{g,t} \leq R_g^{hr} u_{g,t}^m + R_g^{SD}(v_{g,t} - u_{g,t}^m + u_{g,t-1}^m), \forall g \in G, t \in T \quad (7)$$
$$\sum_{q=t-UT_g+1}^{t} v_{g,q} \leq u_{g,t}^m, \forall g \in G, t \geq UT_g \quad (8)$$
$$\sum_{q=t+1}^{t+DT_g} v_{g,q} \leq 1 - u_{g,t}^m, \forall g \in G, t \leq T - DT_g \quad (9)$$
$$v_{g,t} \geq u_{g,t}^m - u_{g,t-1}^m, \forall g \in G, t \in T \quad (10)$$
$$v_{g,t}, u_{g,t}^m \in \{0,1\}, \forall g \in G, t \in T \quad (11)$$

*Network constraints:*
$$P_{k,t} - b_k(\theta_{i,t} - \theta_{o,t}) = 0, \forall k \in K, t \in T \quad (12)$$
$$P_{k,t} = \sum_{n \in N}(PTDF_{k,n}^{ref} P_{g \in g(n),t}), \forall k \in K, t \in T \quad (13)$$
$$-P_k^{max} \leq P_{k,t} \leq P_k^{max}, \forall k \in K, t \in T \quad (14)$$
$$\sum_{g \in g(n)} P_{g,t} + \sum_{k \in \delta^+(n)} P_{k,t} - \sum_{k \in \delta^-(n)} P_{k,t} = d_{n,t}^m, \forall n \in N, t \in T \quad (15)$$
$$\theta_{ref,t} = 0 \; \forall t \in T \quad (16)$$

A SCUC model with relevant base case constraints are formulated to highlight the proposed ST based approach. Based on the above constraints, the SCUC formulation implementing B-θ formulation is represented by (1)-(13) and (15)-(16) whereas, the SCUC formulation implementing PTDF formulation is implemented by (1)-(12) and (14)-(15).

## III. GRAPH GENERATION

In this research we focus on using advanced ML layers, specifically GNN and LSTM. Since GNN is graph-based, this requires each snapshot of the data in graph form. Data from our prior work in [23], was used to curate the samples in graph form. A graph, *M*, is defined as a combination of vertices *V*, edges *E*, and the adjacency matrix *A*, such that *M* ∈ *(V, E, A)*. Graphs are used since power systems are based on topology-based non-Euclidean networks and the network can be defined to specify further relationship for the ML applications. For GNN, vertices are assumed to be the bus nodes, *N*, and Edges are assumed to be transmission lines, *K*.

Additionally, each node and edge can have static features and dynamic features that define their properties. The dynamic features are the features that vary over a time period, *T*, as in the case of forecasted nodal load schedule used in SCUC and is node feature (NF). If a node does not have loads, then the load is assumed to be 0 for 24 hour-period. Using this idea, a node feature matrix, *NF*, is created with dimension of nodes and number of node features *NF = [N, T]*. For output labels associated with nodes, $Y_N$, are formed using the commitment status of generators, $u_{g,t}^m$.

Each line or edge has static edge features (EF) (does not change over a 24-hour period) and these include line reactance, and thermal limits. Since graph theory does not define parallel edges, an equivalent representation of parallel lines is used to denote the respective edge. This implies the line reactance is summed and the minimum thermal limit between parallel lines is used as the feature. This is because if one line in a set of parallel is congested, it implies the other lines are congested in the parallel path. The line/edge matrix, *EF*, consists of a dimension of edges and edge features, *EF = [E, 2]*. Output labels associated with edges, $Y_E$, are formed using the respective line flows, $P_{k,t}$, and line thermal ratings, $P_k^{max}$. If the line is loaded beyond 75%, it is identified as a critical line and the corresponding $Y_E$ is labeled to be 1; otherwise, it is 0.

Additionally, a sparse adjacency matrix, $A$, is required which indicates the relationship of the $E$ and $V$. From the above information, $NF$, $EF$, $A$, and $Y_N$ are utilized to form graphs for node classification whereas $NF$, $EF$, $A$, and $Y_E$ are utilized to form graphs for edge classification using Spektral package [34]. Algorithm 1 describes the graph generation in detail.

| | |
|---|---|
| 1: | **Algorithm 1** ML graph generation |
| 2: | **Repeat** |
| 3: |   **for each** $k \in K$ |
| 4: |     **If** frombus[$k$], tobus[$k$] != frombus[$k-1$], tobus[$k-1$] |
| 5: |       $EF[m, k] = [b_k, P_k^{max}]$ |
| 6: |       $E \cup \{k\}$ |
| 7: |     **Else** |
| 8: |       $EF[m, k-1] = [(b_{k-1} + b_k), min(P_{k-1}^{max}, P_k^{max})]$ |
| 9: |     **End if** |
| 10: |   **for each** $n \in N$ |
| 11: |     **for each** $t \in T$ |
| 12: |       $NF[m, n, t] = d_{n,t}^m$ |
| 13: |       $Y_N[m, n, t] = u_{g(n),t}^m$ |
| 14: |   **for each** $e \in E$ |
| 15: |     **for each** $t \in T$ |
| 16: |       **If** $(P_{e,t} / P_e^{max}) > 0.75$ |
| 17: |         $Y_E[m, e, t] = 1$ |
| 18: |       **Else** |
| 19: |         $Y_E[m, e, t] = 0$ |
| 20: |       **End If** |
| 21: | **until** $m \leq M$ |
| 22: | **for each** $i \in E$ |
| 23: |   **for each** $j \in E$ |
| 24: |     $A[i, j] = 0$ |
| 25: | **for each** $e \in E$ |
| 26: |   $A[frombus[e], tbus[e]] = 1$ |
| 27: | Using Spektral generate $M$ graphs for NC with $NF$, $EF$, $A$ and $Y_N$. |
| 28: | Using Spektral generate $M$ graphs for EC with $NF$, $EF$, $A$ and $Y_E$. |

## IV. PROPOSED OFFLINE APPROACH

### A. ML Approach (Model Reduction)

The trained and developed ML architectures using historical information are utilized to reduce constraints and variables in the SCUC model with high confidence. Specifically, constraints reduction is achieved through EC whereas binary variable reduction is implemented through NC. Both NC and EC rely on the historical data generated as graphs in Section III. The B-θ-SCUC and PTDF-SCUC, described in Section II, where all constraints and variables are utilized, are used as benchmark models for comparison against a ML-R-SCUC process. EC model is trained using graphs which have edge labels to identify each line as critical or non-critical in each period $t$ indicating whether its loading level exceeds a certain threshold 75%. For node classification, the graphs are created with node labels that denote commitment schedule (ON/OFF) of each generator $g$ in each period $t$. The input features in the graphs are identical irrespective of node classification or edge classification, i.e., node features as demand profile in each node and edge features as line properties and thermal limits. Both models provide solutions/predictions for the 24-hour time period. The goal of NC is using ML model to predict the status of generator, g, is ON/OFF in period $t$ in any node with certain confidence (probability).. This is used to reduce binary variables in the SCUC model by fixing respective $u$ and $v$ as constants thereby simplifying the model that needs to be optimized. The goal of EC is to predict/identify the critical and non-critical lines with a certain confidence. This information is utilized to reduce redundant constraints in (14) as non-critical lines are unlikely to breach limits.

As we know, power systems have three key components, namely, network, demand and generators which are utilized for graph generation in Section III. Based on the above logic, the models once trained are utilized in an online setting to optimize the R-SCUC as shown in Fig. 1. It can also be noted that EC or NC can be individually performed as these models are decoupled since two different models are trained.

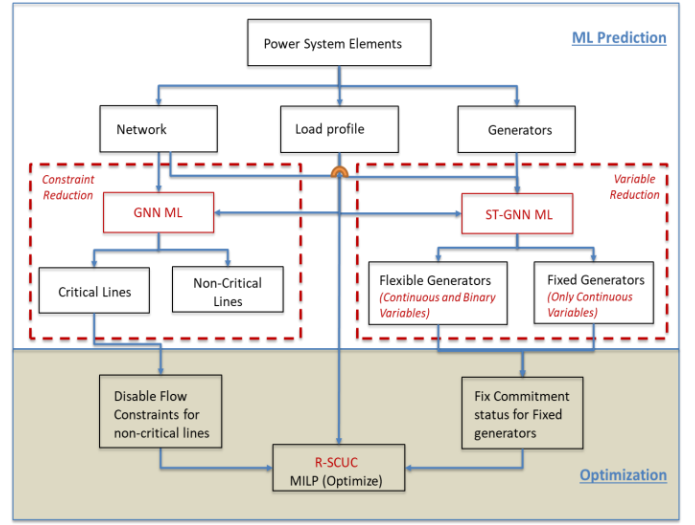

Fig. 1. SCUC model reduction.

### B. GNN Layer

The key component of the ML model considered in this research is inclusion of a GNN layer. GNN differs from traditional DNN models as it considers graph-structured data as opposed to 2-D or flat data. The inclusion of graphs provides an additional benefit that the trained learning model can also considers the topology of the network along with the geographic/spatial understanding of the data. In power systems, the topology of the network is a key property as the power supplied to the demand through pathways or transmission is limited by capability of the network. The network can have bottlenecks which can arise different locational marginal pricing in energy markets (i.e. expensive generators provide power to the demand rather than cheap generators due to network limits). The network has properties such as thermal limits and admittance (inverse electrical distance) that can result in choosing paths with highest admittance and the flow is shared between parallel paths in ratio of admittance (Ohm's law). Provided that such phenomena exist in power system networks, it is paramount that specialized models are required which have the ability to learn network properties. Therefore, GNN uses graph-structured inputs which depict the power network. In comparison, the DNN architecture assumes such relationship can be realized during training. This may not achieve top performance and very high accuracy.

In GNN, it can also be noted that having additional layers can also benefit the process (though computationally training expensive) as each node obtains information from more neighboring nodes and nodes connected by the edges at a specific node (more layers imply more hops from a node). In this work, two specific layers were considered from the Spektral package, namely, edge conditioned convolution (ECC) [35] and XENET [36]. The difference between the two relies in the outputs at a layer level.

The ECC layer only provides activated outputs for each node $i$ as $x_i^{'}$ in (17). However, while identifying the activated output for each node $i$, the ECC layer also utilized information from neighboring nodes. This is implemented for each neighboring node $j$ that is connected to node $i$ through edge $e_{j,i}$. The MLP is a multi-layer perceptron that outputs an edge-specific weight as a function of edge attributes. By considering neighboring nodes, ECC can benefit NC further since edge specific weights are summed at a nodal level [35]. However, due to this, ECC can only be utilized for NC.

$$x_i' = x_i W_{root} + \sum_{j \in N(i)} x_j MLP(e_{j,i}) + b \quad (17)$$

In comparison, a XENET layer computes pre-activated values using (18)-(20) for each node $i$. Using the pre-activated values, XENET layer finally provides activated values for node $i$ as $x_i^{'}$ and each edge connected by nodes $i$ and $j$ as $e_{i,j}^{'}$ in (21) and (22), respectively. Since XENET layer provides activated values for both nodes and edges, it can be utilized for both NC or EC.

$$s_{i,j} = PRELU((x_i||x_j||e_{i,j}||e_{j,i})W^{(s)} + b^{(s)}) \quad (18)$$
$$s_i^{(out)} = \sum_{j \in N(i)} s_{i,j} \quad (19)$$
$$s_i^{(in)} = \sum_{j \in N(i)} s_{j,i} \quad (20)$$
$$x_i' = \sigma_g((x_i||s_i^{(out)}||s_i^{(in)})W^{(n)} + b^{(n)}) \quad (21)$$
$$e_{i,j}' = \sigma_g(s_{i,j}W^{(c)} + b^{(c)}) \quad (22)$$

### C. LSTM Layer

The second specialized layer proposed in this research is an LSTM layer. The LSTM layer is a form of RNN and performs well on temporal data and it takes sequential data as inputs. An LSTM layer consists of several LSTM cells for each time period. Each cell in the LSTM layer is for time-period $t$ with input $x$, to provide an output state $o$, a cell state $c$, and a hidden state $h$, while considering a memory state $h$ and a cell state $c$ from period $t-1$. These values are calculated using (23)-(28). Fig. 2 shows LSTM layer rolled out for 2 hours. Since the problem considered in this paper is SCUC, each sample consists of 24-hour period of data and therefore the resulting layer will have 24 cells. The result from the final time period is obtained through the output state $o$, when rolled out across 24-hour time period.

$$f_t = \sigma_g(W_f x_t + U_f h_{t-1} + b_f) \quad (23)$$
$$i_t = \sigma_g(W_i x_t + U_i h_{t-1} + b_i) \quad (24)$$
$$o_t = \sigma_g(W_o x_t + U_o h_{t-1} + b_o) \quad (25)$$
$$c_t' = \sigma_c(W_c x_t + U_c h_{t-1} + b_c) \quad (26)$$
$$c_t = f_t c_{t-1} + i_t c_t' \quad (27)$$
$$h_t = o_t \, \sigma_g(c_t) \quad (28)$$

where $b_f$, $b_i$, $b_o$, $b_c$ are non-time dependent biases; $W_f, W_i, W_o, W_c, U_f, U_i, U_o, U_c$ are non-time dependent weight matrices; $f_t$ is forget gate; $i_t$ is input gate; $o_t$ is output gate; $c_t$ is cell state; $h_t$ is hidden state and $x_t$ is the input. $\sigma_g$ is a sigmoid activation function and $\sigma_c$ is a hyperbolic tangent (tanh) activation function.

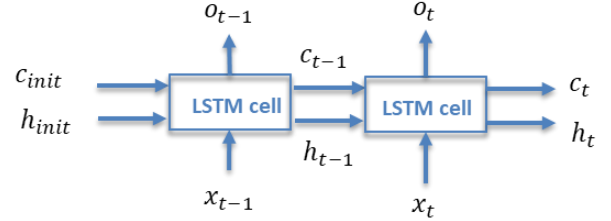

Fig. 2. A 2-hour LSTM layer.

### D. ST Approach

Since the demand and generators in the system is geographically distributed, a spatial correlation of the data is required to be studied. This is considered using GNN layer where the inputs are represented in the form of graphs which mimic the network structure of power system. In addition, the temporal correlation of the data can also be studied since SCUC is an optimization problem for a day-ahead 24-hour period. Therefore, relationships between various hours of the day can be studied using the LSTM layer. By utilizing both the above-mentioned layers, the ML model takes into consideration a ST relationship between inputs and outputs. A ST approach is obtained if the outputs of GNN layer is passed through to LSTM layer and finally the 24-hour prediction is obtained as an output. The ML model is then trained in batches to form this relationship with the training graphs and evaluated with the testing graphs. A validation data graph set is used to aid the training to avoid overfitting the model.

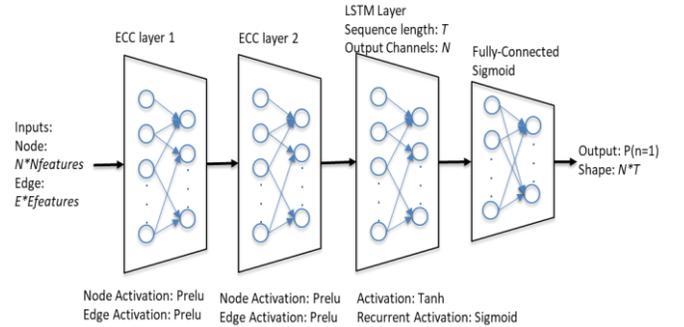

Fig. 3. Spatio-temporal ML architecture for node classification.

NC is implemented with a ST model to predict the commitment status of generators for any given graph input for a 24-hour period. For generators, LSTM is important as there are several temporal constraints like ramp restrictions and min-off/on time limits which are coupled to the commitment schedule. The GNN layers and LSTM layers together form an advanced ML architecture for ST. In Fig. 3, a node-classification ST model is trained to fit inputs of load profile and respective commitment labels where the input is fed through multiple GNN layers with activation and the resulting node embedding is then passed as inputs for the LSTM layer. Multiple GNN layers are utilized since each layer tries to identify the nodal relationship of neighboring nodes and additional layers gather information of nodes multiple hops from each node.

However, it was noted that the line criticality status predicted in EC does not vary as much as generator commitment schedule and only a few bottleneck lines are critical or congested in the network for majority of the time. Therefore, EC has a less temporal correlation but a high spatial correlation between the inputs and outputs. In EC, LSTM layers results in poor prediction due to the above reason, whereas GNN layer independently outperforms an ST model. In Fig. 4, edge-classification ST model is trained to fit inputs of nodal load profile and critical lines in the system. Once the models are trained, the ML predictions are utilized not only to reduce binary variables but also reduce redundant constraints to bring about additional time-savings by further reducing the SCUC.

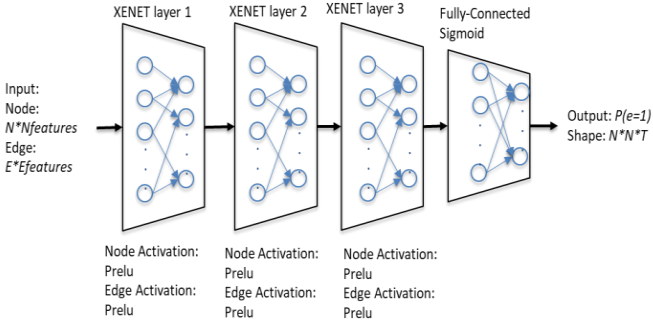

Fig. 4. Spatial ML architecture for edge classification.

In both NC and EC, a final fully-connected sigmoid activation layer is utilized to obtain the probability of the output being 1. In the case of NC, it is probability of a generator in a particular node being ON. In the case of EC, it is the probability of the line being critical by breaching threshold of 75% of the line thermal limit.

### E. Training

Both NC and EC are trained and tested using the respective graphs generated described in Section III. Only feasible samples of the SCUC are considered for graph creation and a total of 1800 graphs/samples denoted as $M$ are created for each test system. The $M$ graphs/samples are shuffled and split into three datasets: 70% training samples (1260 samples) denoted as $M^{train}$, 15% validation samples (270 samples) denoted as $M^{val}$, and 15% testing samples (270 samples) denoted as $M^{test}$. The respective models are trained using training samples $M^{train}$, in epochs. In each epoch, the weights of the layers are tuned to fit the graphs in mini-batches of 64 to train the model whereas the validation dataset, $M^{val}$, is shuffled and utilized to identify the best parameters to fit the model to avoid overfitting. Finally, the trained models are tested on testing samples, $M^{test}$, to see if the ML model has generalized to predict.

### F. Verification

The output/predictions for the ML models are probabilities as described in sub-section IV-D. The predictions describe a confidence measure through probabilities. Therefore, a post-processing/classification of the predictions is performed to utilize in model reduction during verification on test-samples, $M^{test}$. For NC, the probability of generator $g$ being ON at a particular time-period $t$ for any sample $m$, $P(u_{g,t}^m)$, is used to reduce the number of binary variables associated with generator commitment schedule, $u_{g,t}^m$, in both B-θ-SCUC and PTDF-SCUC. The following principle is utilized for variable-reduction in NC for $m \in M^{test}$:

- Reduce variable, $fix\ u_{g,t}^m = 1$ if $P(u_{g,t}^m) \geq 0.90$.
- Reduce variable, $fix\ u_{g,t}^m = 0$ if $P(u_{g,t}^m) \leq 0.10$.
- Warm-start, $let\ u_{g,t}^m = 0$ if $0.10 < P(u_{g,t}^m) < 0.50$.
- Warm-start, $let\ u_{g,t}^m = 1$ if $0.5 \leq P(u_{g,t}^m) < 0.90$.

For EC, the probability of line $k$ being critical in any time-period $t$ for any sample $m$, $P(l_{k,t}^m)$, is utilized to determine if thermal limit constraints (14) are active in both B-θ-SCUC and PTDF-SCUC. The following principle is utilized for constraint-reduction for $m \in M^{test}$:

- Make constraint (14) inactive for any line $k$, if $P(l_{k,t}^m) < 0.50$ for all $t \in T$.
- Make constraint (14) active for any line $k$, if $P(l_{k,t}^m) \geq 0.50$ for any $t \in T$.

The above-described principles can be utilized to obtain the respective R-SCUC model through variable-reduction in NC model and constraint-reduction in EC for both B-θ-SCUC and PTDF-SCUC. It can be noted that separate ML models exist for NC and EC, therefore, they can be utilized either together or separately. If only constraint-reduction is implemented, then it is represented as constraint-R-SCUC (C-R-SCUC). If only variable-reduction is implemented, then it is represented as variable-R-SCUC (V-R-SCUC). If both variables and constraints are reduced, then it is termed as variable and constraint R-SCUC (VC-R-SCUC).

### G. Confidence Measurement

The model accuracy can be verified using the classified outputs. Once the model is trained, the model accuracy can be calculated for training dataset $M^{train}$ and testing dataset $M^{test}$. The classified outputs based on the predicted output probabilities from NC and EC are utilized for accuracy calculation. For NC this is implemented for $t \in T$ and $g \in G$, if $P(u_{g,t}^m) \geq 0.5$ is classified as 1/ON and $P < 0.5$ as 0/OFF. For EC, $P(l_{k,t}^m) \geq 0.5$ is classified as 1/critical and $P(l_{k,t}^m) < 0.5$ as 0/non-critical. The overall accuracy can be calculated using (29).

$$Acc = 1 - \frac{1}{N_m * N_i * N_t} \sum_{m=1}^{N_m} (\sum_{i \in I} \sum_{t \in T} |y_{m,i,t} - y'_{m,i,t}|) \quad (29)$$

where $N_m$, $N_i$, $N_t$ represent the number of samples, number of resource and number of time-periods, respectively; $y$ represents the ML classified predictions and $y'$ is the true value/label. For NC, resource are generators $g \in G$, whereas for EC, the resources are $k \in K$.

Model accuracy provides a metric of model generalization. However, the testing samples $M^{test}$ are verified in an online setting by using the trained model. Therefore, additional metrics namely, base-normalized cost (BNC) and base-normalized time saved (BNTS), are defined in (30) and (31). Both BNC and BNTS are percent values of the base values. The non-reduced optimization results for SCUC are used as base values for the respective formulations and compared to the proposed C-R-SCUC, V-R-SCUC, and VC-R-SCUC models. Consistency is maintained by calculating BNC and BNTS separately for B-θ and PTDF formulations.

$$\text{BNC} = \left(\frac{|Base\ Cost - Reduced\ model\ Cost|}{Base\ Cost}\right) * 100\% \quad (30)$$

$$\text{BNTS} = \left(\frac{|Base\ Solve\ Time - Reduced\ Model\ Solve\ Time|}{Base\ Solve\ Time}\right) * 100\% \quad (31)$$

## V. RESULTS AND ANALYSIS

For ML step, it is implemented in Python 3.6 using spectral, keras, and tensorflow packages. The SCUC model is implemented in Python using Pyomo [37]. The data creation and verification steps are implemented using AMPL and solved using Gurobi solver. A computer with Intel® Xeon(R) W-2295 CPU @ 3.00GHz, 256 GB of RAM and NVIDIA Quadro RTX 8000, 48GB GPU was utilized.

TABLE I. TEST SYSTEM SUMMARY

| System | Gen cap (MW) | # bus | #gen | # branch |
|---|---|---|---|---|
| IEEE 24-Bus [38] | 3,393 | 24 | 17 | 38 |
| IEEE 73-Bus [38] | 10,215 | 73 | 45 | 117 |
| IEEE 118-Bus [39] | 5,859 | 118 | 54 | 186 |
| South Carolina (SC) [40] | 12,189 | 500 | 90 | 597 |

The standard test systems summarized in Table I were utilized for simulations and results analysis. For optimization, MIPGAP=0.001 is used consistently for all verification/online results.

### A. NC Training Summary

Initially the results of ST model were compared for day ahead NC alone (i.e. generator commitment prediction). This is because the benchmark used is a DNN model proposed in [23] which only classifies generator commitment schedules. The goal of this section is to identify the benefits of using advanced ML architectures with GNN and LSTM layers in terms of prediction accuracy.

From Table II, we notice that for both training and testing accuracy for NC. Here, ST-ML model brings about significant improvements in predictions as denoted by the accuracy. Typically, the proposed ST model outperforms the benchmark DNN by 0.75%–1.6% in testing accuracy across all the test systems. The accuracy was calculated using (29) as described in sub-section IV.G.

It can be noted that both models work on similar data inputs in terms of demand schedule to predict commitment schedule. But the key distinction is that ST model uses GNN layer with graph inputs including topology features and LSTM layer for temporal correlation which brings the performance benefits.

TABLE II. MODEL ACCURACY FOR NODE CLASSIFICATION

| #Bus | DNN Accuracy (%) | | ST Accuracy (%) | |
|---|---|---|---|---|
| | Training | Testing | Training | Testing |
| 24 | 97.16 | 97.01 | 98.40 (↑ 1.24) | 98.31 (↑ 1.30) |
| 73 | 95.82 | 95.65 | 97.24 (↑ 1.42) | 97.04 (↑ 1.39) |
| 118 | 97.83 | 97.62 | 98.99 (↑ 1.16) | 98.96 (↑ 1.34) |
| 500 | 99.06 | 99.04 | 99.80 (↑ 0.76) | 99.79 (↑ 0.75) |

The learning curve for ST model in Fig. 5 shows both the training and validation curves based on the probabilistic/pre-classification outputs. The test system with the worst accuracy, IEEE 73-bus system, was used to present this result. The training was performed in batches and in each epoch, the shuffled validation dataset is utilized to obtain the best training score. Fig. 6 represents the model loss as a function of training epochs and it can be noted that the ST-ML model generalizes well over the validation samples without overfit or underfit.

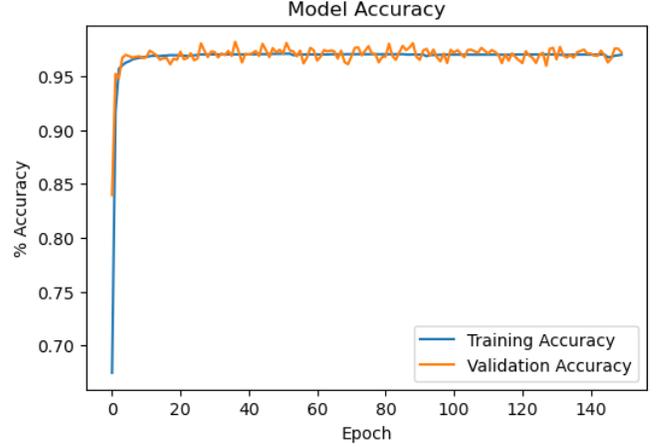

Fig. 5. Training curves for IEEE 73-bus system for NC with ST.

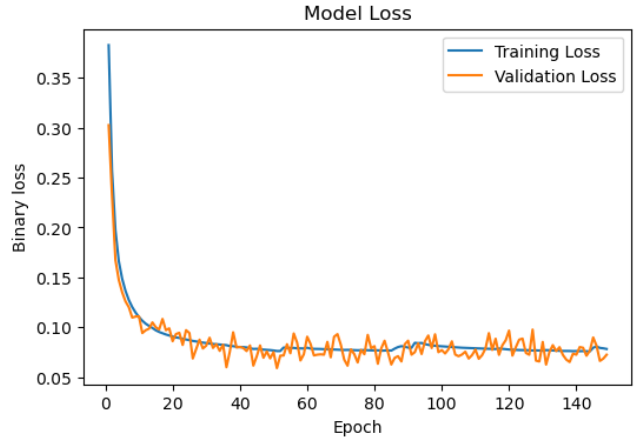

Fig. 6. Model loss curves for IEEE 73-bus system for NC with ST.

Additionally, Fig. 7 and Fig. 8 show the histogram of predictions errors in the test samples of the IEEE 73-bus system for ST and DNN ML models, respectively. The trained models were utilized to obtain the post-classification outputs on the testing samples (not used in training). For each testing sample, the post-classification outputs were validated for the number of wrong predictions with respect to respective true values. The IEEE 73-bus system was utilized as this system has the lowest accuracy. For NC, it has 45 generators spread across 73 nodes and makes a total of 1080 predictions over the 24-hour time-period for each $m \in M^{test}$. Due to the nature of ML, all samples may have a few prediction errors when classification is performed. However, we note that ST model results in fewer prediction errors in comparison to DNN model.

Since ML can result in errors, we need to be careful to utilize the ML prediction solutions. Therefore, the outputs are used for selective SCUC model reduction, i.e., utilize the outputs with high confidence for model reductions in R-SCUC. It can be noted that 1 wrong prediction can result to model infeasibilities in SCUC as these variables are temporally correlated and also require to satisfy reliability requirements in (3)-(11).

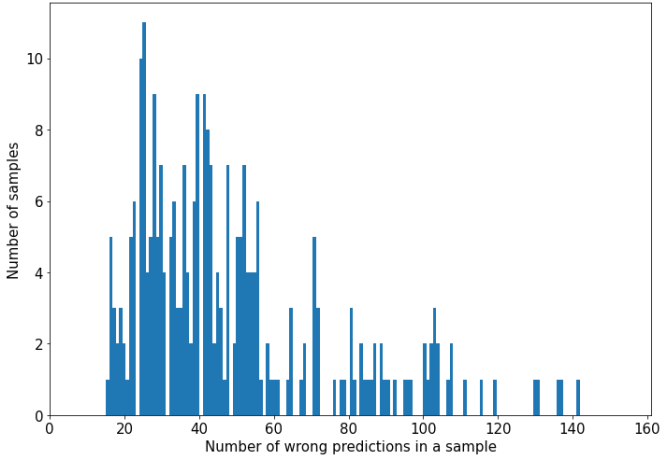
Fig. 7. Histogram of ST ML predictions on $M^{test}$ for IEEE 73-bus system.

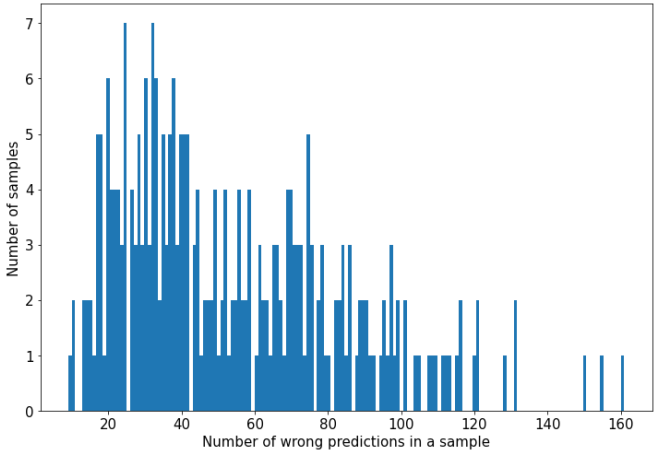
Fig. 8. Histogram of DNN predictions on $M^{test}$ for IEEE 73-bus system.

*B. NC Verification Benchmark Comparison*

A model verification is performed as described in sub-section IV.F and the metrics, BNC and BNTS, defined in sub-section IV.G are utilized to compare the ST and DNN predictions in an online setting. Here, the B-θ formulation of SCUC is used as a base method and DNN R-SCUC is used as the benchmark method against ST R-SCUC. It can be noted once again that the results presented only include NC as the benchmark model, DNN R-SCUC, in [23] only performs variable reduction. Both the benchmark model and ST NC model are verified using the procedure detailed in sub-section IV.F to reduce and warm-start commitment variables. The procedure and formulation are kept consistent to make the comparison unbiased.

We observed that the benchmark method, DNN-R-SCUC, results in a few infeasible samples due to larger wrong predictions whereas ST-R-SCUC does not result in any infeasibilities across all testing samples. Additionally, it was noted that on average across all test systems, ST-R-SCUC resulted in 38.6% time saved whereas the feasible test samples of DNN-R-SCUC resulted only in 32.03% computational savings over SCUC. This trend is because ST NC model reduces larger number of variables in R-SCUC accurately compared to DNN model. Furthermore, the computational savings are more consistent across all test system using ST whereas DNN results in inconsistent computational savings. Not only that, on average across all test systems, the solution quality of 0.04% was noted which is much lower than the MIPGAP for ST-R-SCUC and betters the solution quality obtained from DNN-R-SCUC. This implies that DNN model, a trade-off between solution quality and computational time savings are noted which requires a control for post-processing to be effectively tuned for each test system whereas the ST model outperforms DNN and generalizes well irrespective of the test system.

TABLE III. MODEL ACCURACY FOR NODE CLASSIFICATION

| System | 24-Bus | | 73-Bus | | 118-Bus | | 500-Bus | |
|---|---|---|---|---|---|---|---|---|
| Model | DNN | ST | DNN | ST | DNN | ST | DNN | ST |
| Infeasible samples | 0 | 0 | 7 | 0 | 4 | 0 | 13 | 0 |
| BNC (%) | 0.00 | 0.02 | 0.12 | 0.11 | 0.30 | 0.00 | 0.13 | 0.04 |
| BNTS (%) | 3.92 | 24.8 | 49.2 | 44.2 | 38.7 | 39.7 | 36.3 | 45.5 |

*C. Verification Results for Proposed Methods*

For this section, the effects of NC for variable reduction represented as V-R-SCUC, EC for constraint reduction, represented as C-R-SCUC and the combination of both NC and EC represented as VC-R-SCUC were studied for model reduction. Also, it can be noted that the three approaches were verified for R-SCUC for both B-θ formulation and PTDF formulation of SCUC. Sub-section V.B showed that ST model outperforms DNN models for variable reduction, but constraint reduction was not studied. For each test system, variable reduction and/or constraint reduction are performed by first obtaining an offline trained model. The trained models are verified in an online setting for test samples, $m \in M^{test}$. For B-θ R-SCUC, the base is considered as B-θ SCUC whereas for PTDF R-SCUC, the base is considered as PTDF SCUC, which is to maintain consistency and avoid errors.

In general, from Table IV and Table V, we notice a general pattern that C-R-SCUC (constraint reduction) yields lower computational time savings than V-R-SCUC (binary variable reduction) whereas the combination VC-R-SCUC yields the highest time savings. This is because reducing binary variables by fixing commitment schedules will reduce problem complexity more than removing redundant constraints in the problem. From Table IV, on average across all test systems and test samples, we can see that B-θ R-SCUC results in a computational time savings of 18.7% for C-R-SCUC, 38.6% for V-R-SCUC and 53.9% for VC-R-SCUC. Similarly, from Table V, on average across all test systems and test samples, we notice PTDF R-SCUC results in computational time savings of 21.6% for C-R-SCUC, 41.17% for V-R-SCUC and 56.2% for VC-R-SCUC. On average across both formulations and test system, it was noticed that C-R-SCUC resulted in time savings of 20.2% whereas V-R-SCUC and VC-R-SCUC resulted in 39.9% and 55.1%, respectively. Therefore, VC-R-SCUC can provide significant time savings benefiting from reducing binary variables and removing redundant constraints to reduce the problem complexity as seen from the above results. This is achieved without compromising the solution quality.

The advantage with C-R-SCUC is that only non-critical constraints are removed and therefore this results in no change in solution quality with respect to non-reduced SCUC model since the BNC is 0.00%. This implies that R-SCUC model results in the same solution as the SCUC problem despite removing constraints. On the other hand, variable reductions are associated with minor change in solution quality in R-SCUC when compared with SCUC. However, this change is on average ~0.04% for both B-$\theta$ R-SCUC and PTDF R-SCUC. This change is lower than the MIPGAP considered in solving the problem which is 0.1%. By combining both EC and NC, the solution quality difference only results from NC since EC only removes redundant constraints.

It was noted that PTDF formulation is faster than B-$\theta$ formulation due to inherent reduced constraints and variables in (13) as opposed to (12), which results in faster computation; however the solution quality between the two only depends on the accuracy of PTDF matrix formed. A static topology is used in these results. B-$\theta$ formulation is easier to handle changes in topology but PTDF formulation requires recalculating the matrix for any changes in topography which will be considered in future research.

TABLE IV. VERIFICATION RESULTS FOR B-$\theta$ R-SCUC

|  | C-R-SCUC | | V-R-SCUC | | VC-R-SCUC | |
|---|---|---|---|---|---|---|
| Metric | BNC (%) | BNTS (%) | BNC (%) | BNTS (%) | BNC (%) | BNTS (%) |
| 24-Bus | 0.00 | 18.22 | 0.02 | 24.76 | 0.02 | 44.72 |
| 73-Bus | 0.00 | 28.22 | 0.11 | 44.23 | 0.11 | 56.50 |
| 118-Bus | 0.00 | 10.52 | 0.00 | 39.72 | 0.00 | 57.44 |
| 500-Bus | 0.00 | 17.73 | 0.04 | 45.51 | 0.04 | 56.92 |

TABLE V. VERIFICATION RESULTS FOR PTDF R-SCUC

|  | C-R-SCUC | | V-R-SCUC | | VC-R-SCUC | |
|---|---|---|---|---|---|---|
| Metric | BNC (%) | BNTS (%) | BNC (%) | BNTS (%) | BNC (%) | BNTS (%) |
| 24-Bus | 0.00 | 12.50 | 0.02 | 27.00 | 0.02 | 35.62 |
| 73-Bus | 0.00 | 30.78 | 0.12 | 40.46 | 0.12 | 61.85 |
| 118-Bus | 0.00 | 19.95 | 0.00 | 53.33 | 0.00 | 67.85 |
| 500-Bus | 0.00 | 23.17 | 0.03 | 43.87 | 0.03 | 59.48 |

## VI. CONCLUSIONS

In this paper, an advanced ST model utilizing GNN and LSTM layer to study the spatial correlation and temporal correlation, respectively, of SCUC was proposed. The use of GNN requires data to be formatted to a graph form and a procedure to obtain input graphs was developed. In particular, the proposed ML model was utilized for variable reduction and/or constraint reductions in an online setting by utilizing the ML predictions. It was seen that variable reduction and constraint reduction are decoupled and can be implemented separately or jointly. The proposed C-R-SCUC, V-R-SCUC and VC-R-SCUC approaches present significant time savings without loss in solution quality. In addition, the proposed V-R-SCUC, C-R-SCUC and VC-R-SCUC are formulation agnostic and was shown to be very effective on different SCUC formulations namely, B-$\theta$ SCUC and PTDF SCUC. It can also be implemented in existing formulations, decomposed, heuristic approaches for SCUC without issues since ML models training and predictions are de-coupled to online implementation in R-SCUC.

Furthermore, the benchmark system, DNN for V-R-SCUC from [23] was used for comparing the advantages of the ST model. It was seen that ST model outperforms the state-of-the-art in both solution quality and computational time savings. In particular, ST approach results in better predictions which results in a high solution quality with the absence of any infeasibility problems. In [23], DNN required additional feasibility layers to post-process the ML prediction in an online setting which was eliminated in ST. In addition to model accuracy, ST V-R-SCUC results in consistent time savings across all test systems and outperforms DNN R-SCUC.

Additional time savings were introduced by considering constraint reduction and enabling VC-R-SCUC. Overall, on average across various SCUC formulations, test systems and test samples, it was noticed that VC-R-SCUC results in significant time savings of 55.1% when compared to SCUC implementation. This was much better than C-R-SCUC and V-R-SCUC which results in 20.2% and 39.9% time-savings, respectively, with respect to SCUC. This is important since most prior research either implements variable reduction or constraint reduction but not a combination of both.